\documentclass{article}

\usepackage{arxiv}

\usepackage[utf8]{inputenc} 
\usepackage[T1]{fontenc}    
\usepackage{hyperref}       
\usepackage{url}            
\usepackage{booktabs}       
\usepackage{amsfonts}       
\usepackage{nicefrac}       
\usepackage{microtype}      
\usepackage{lipsum}		
\usepackage{graphicx}
\usepackage{natbib}
\usepackage{doi}
\usepackage{amsmath}
\usepackage{algorithm}
\usepackage{algpseudocode}
\usepackage{amssymb}
\usepackage{multirow}
\usepackage{hyperref}
\usepackage[table]{xcolor}
\definecolor{highlight}{gray}{0.92}
\let\cite\citep

\title{REGEN: Replay-recycling for Expert-to-Generalist distillation with Offline Reinforcement Learning}

\author{
Yunjie Chen$^{1,2,3}$,\quad
Zihao Chen$^{1,3}$, \quad
Xiaoxin Chen$^{1}$,\quad
Fan Wang$^{1,2}$\thanks{Corresponding to: fanwang.px@gmail.com}\\[0.6em]
$^{1}$vivo AI Lab, Shenzhen, China\\
$^{2}$Qing Yan Technology, Shenzhen, China \\ 
$^{3}$Sun Yat-Sen University, Guangzhou, China\\[0.3em]
}

\begin{document}
\maketitle

\begin{abstract}
Large-scale online reinforcement learning (RL) is the predominant means of eliciting advanced abilities including long-term reasoning and agentic tool use in large language models (LLMs). 
However, continuing to scale it across vast task domains of interest remains challenging in both computational infrastructure and cost, especially when considering RL as merely a one-off learning stage. 
Recently, a widely used technique for distilling knowledge across various domains and training stages, multi-teacher on-policy distillation (MOPD), helps to decouple the RL stage, saving costs, while maintaining generality across vast domains. 
Nonetheless, similar to online RL, MOPD requires coupled inference and backward passes, which continue to limit its scalability and computational efficiency. 
To address these challenges, we propose \textbf{REGEN: Replay-recycling for Expert-to-Generalist distillation with Offline RL}. Instead of distilling from multiple teacher models, REGEN trains a generalist by simply recycling the replay memory—the free by-product of the teachers' specialized RL training—and employing offline RL algorithms. REGEN completely decouples the rollout sampling from the backward training process and thus greatly reduces the training cost.
Across mathematical reasoning, code generation, and instruction following, REGEN matches the accuracy of MOPD at substantially lower cost. 
It potentially turns online RL into a data synthesis process instead of a one-off learning stage, and can be extended to large-scale post-training without requiring heavy computational load. Code is available at \url{https://github.com/yunjie-sysu/REGEN}
\end{abstract}


\section{Introduction}
\label{sec:intro}


Reinforcement learning (RL) has emerged as the dominant means of endowing large language models (LLMs) with strong reasoning and decision-making ability, underpinning recent advances in mathematical problem solving, program synthesis, scientific question answering, and agentic tool use~\cite{DBLP:journals/corr/abs-2501-12948,DBLP:journals/corr/abs-2412-16720}. However, as the field shifts from optimizing models for individual tasks toward training generalist reasoning agents, large-scale online RL is approaching a structural bottleneck due to the following reasons.
First, jointly optimizing a single policy across many task domains can be relatively challenging and expensive owing to the phenomenon of negative transfer~\cite{yu2020gradient,ahn2025prevalence,wu2026imbalanced}, especially when it comes to RL, which requires complex infrastructure and substantial exploration costs, or even different reward models and hyper-parameters across different domains.
Second, online RL is inherently a one-off optimization process that does not lend itself to straightforward aggregation across domains or stages. Historical rollouts are seldom reused in subsequent training runs—even when the training tasks remain unchanged—leading to a significant waste of computational resources whenever base models or task distributions are switched.

To mitigate the cost of large-scale RL across diverse domains and training stages, multi-teacher distillation~\cite{DBLP:journals/corr/HintonVD15,DBLP:conf/kdd/YouX0T17} has become a practical alternative, in which multiple domain-specialized RL experts are first trained and then serve as teachers for a separate distillation stage. A central obstacle in distilling RL teachers is the well-known problem of \emph{distribution shift}~\cite{DBLP:journals/jmlr/RossB10}, whereby the trajectory distribution induced by the student diverges from that of the teacher. To address this issue, data aggregation~\cite{DBLP:journals/jmlr/RossB10} and on-policy distillation (OPD)~\cite{DBLP:conf/iclr/AgarwalVZSGGB24,DBLP:conf/iclr/Gu0WH24,DBLP:conf/emnlp/KimR16} supervise the student on its own generated trajectories using teacher-provided signals. In the multi-domain setting, the Multi-teacher OPD (MOPD)~\cite{DBLP:journals/corr/abs-2601-02780,ma2026mopd} routes each prompt to the corresponding domain-expert teacher, and thus serves as the natural distillation baseline. Although these methods effectively alleviate distribution shift, they require online inference, supervision, and backpropagation to proceed simultaneously, as in RL, and consequently remain computationally expensive.

\begin{figure}
    \centering
    \includegraphics[width=\linewidth]{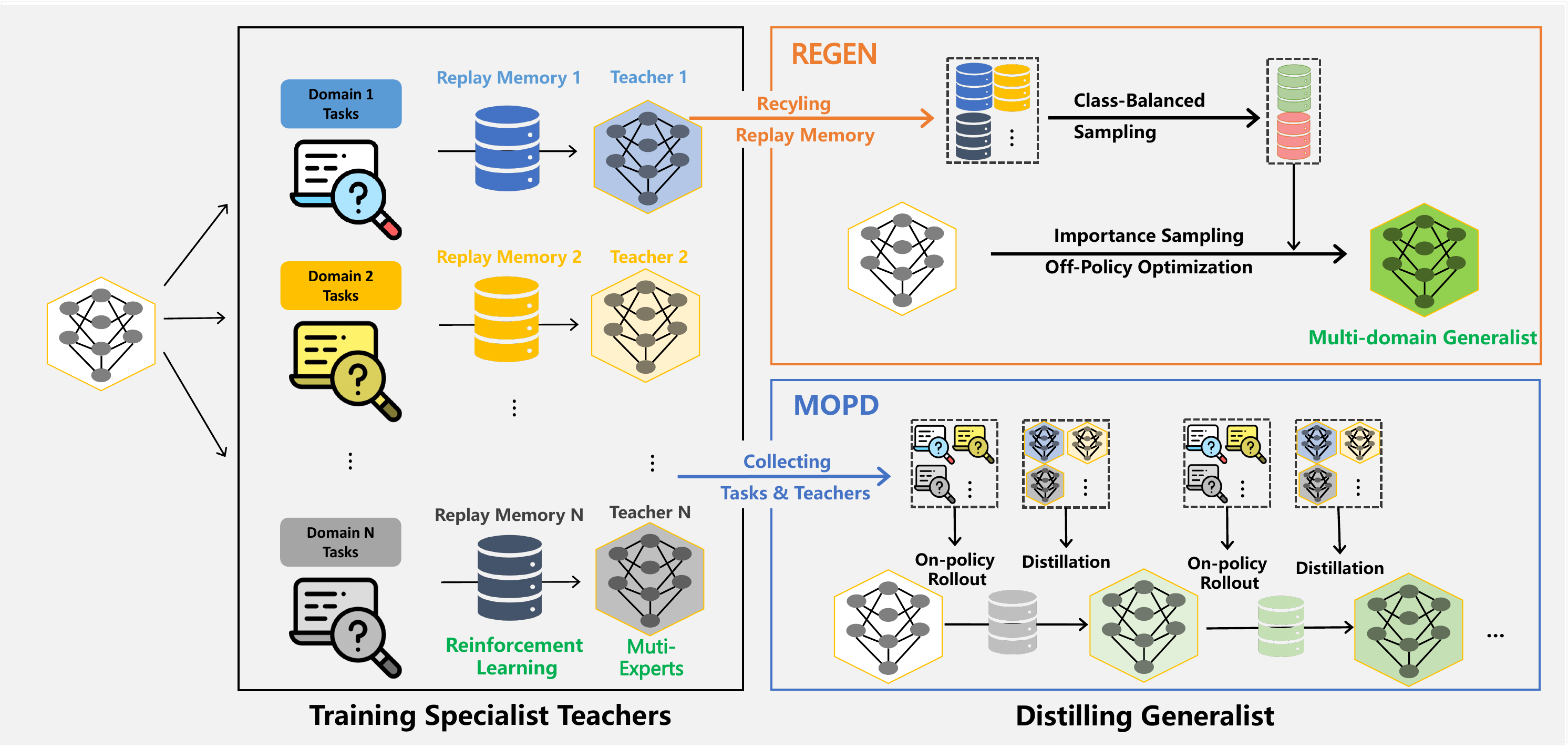}
    \caption{Sketch of the Replay-recycling for Expert-to-Generalist Distillation with Offline RL (REGEN) method in comparison with Multi-Teacher On-Policy Distillation (MOPD)}
    \label{fig:regen-framework}
\end{figure}

Motivated by the observation that the rollout trajectories generated as a by-product of an online RL run  constitute a rich source of supervision typically discarded by existing pipelines, we propose \textbf{REGEN}: \textbf{R}eplay-recycling for \textbf{E}xpert-to-\textbf{GEN}eralist Distillation with Offline RL as an alternative to MOPD. As shown in Figure~\ref{fig:regen-framework}, instead of distilling from trained teachers, REGEN aggregates the replay memory collected during specialist training into a single offline dataset, on which a generalist student is optimized. This method fully decouples data generation from student training, eliminating the need for teachers or rollout engines in the training loop. Through an appropriate combination of a data sampling and an off-policy offline RL objective, REGEN achieves performance and robustness comparable to MOPD at substantially lower computational cost. More broadly, it reframes any RL training run as a data-production pipeline, offering a scalable path toward large-scale RL on large models, including LLMs and beyond.

To summarize, our contributions are threefold. First, we propose \textbf{REGEN}, an offline post-training framework that reuses the replay memories generated during RL teacher training. By recycling these trajectories and optimizing the generalist with offline RL, REGEN achieves performance close to MOPD at substantially lower computational cost, as demonstrated in Section~\ref{sec:exp:main}. Second, Section~\ref{sec:exp:ablation} shows that REGEN provides significantly stronger diversity exploration than MOPD, yielding superior performance in terms of pass@k. Third, REGEN demonstrates the value of replay memory as a reusable asset for LLM post-training, suggesting a new direction in which existing RL runs can serve not only as optimization procedures but also as data-generation processes for subsequent training stages.

\section{Related Work}
\label{sec:related}
\subsection{Reinforcement Learning for LLM Post-Training}
Building on the vast abilities and knowledge acquired from pre-training and supervised fine-tuning (SFT)~\cite{ouyang2022training}, RL adapts the LLM to human preferences~\cite{ouyang2022training,DBLP:conf/nips/ChristianoLBMLA17,DBLP:conf/nips/Ouyang0JAWMZASR22} and other more specific objectives and advanced abilities, including mathematical and programming tasks~\cite{DBLP:journals/corr/abs-2501-12948}, chain-of-thought reasoning~\cite{guo2025deepseek}, and agentic tool use~\cite{zhang2026the}.
Commonly used RL algorithms for LLMs range from proximal policy optimization (PPO)~\cite{DBLP:journals/corr/SchulmanWDRK17}, to group relative policy optimization (GRPO)~\cite{DBLP:journals/corr/abs-2402-03300}, most of which depend on on-policy optimization, where the optimized model continually generates rollouts for training itself in the next stage.
A constraint in on-policy RL methods is that the distribution gap between the trajectories used for training and the current policy cannot be too large. This requires RL to proceed through consecutive inference, evaluation, and training. This coupled process raises considerable challenges for infrastructure and learning in several perspectives: First, RL can be much less sample efficient, as it spends most of its costs on exploring relatively low-valued trajectories. Second, considerable efforts are required for the synchronization between inference, evaluation, and training in order to maximize the utility of the computational resources~\cite{sheng2025hybridflow,hu2024openrlhf}. Third, the modern RL training pipelines for LLMs across different domains can differ a lot in environment settings, including evaluation methods, raising additional challenges for a unified training infrastructure.

\subsection{Policy Distillation from Multiple Teachers}
To address the bottlenecks of building a unified RL framework for all domains, divide-and-conquer RL training combined with policy distillation from multiple teachers offers an alternative path with lower complexity and greater flexibility. LLMs are first trained separately in multiple domains; these domain specialists are then distilled into a generalist capable of performing across domains. Policy distillation is a long-standing topic, with Behaviour Cloning (BC)~\cite{DBLP:conf/nips/Pomerleau88} constituting its most direct realization, formulating policy learning as supervised maximum-likelihood estimation over expert trajectories~\cite{DBLP:conf/nips/Pomerleau88}. A well-documented limitation of BC is \emph{distribution shift}: since the policy is trained under the state distribution induced by the expert but deployed under its own distribution, per-step prediction errors accumulate over the horizon and degrade long-sequence performance~\cite{DBLP:journals/jmlr/RossB10}. This challenge has been addressed by shifting the student's state distribution, leading to approaches such as Data Aggregation (DAgger)~\cite{DBLP:journals/jmlr/RossGB11} and inverse RL~\cite{ng2000algorithms}. When it comes to optimizing LLMs, BC and DAgger take the form of knowledge distillation~\cite{DBLP:journals/corr/HintonVD15} and on-policy distillation (OPD)~\cite{DBLP:conf/iclr/AgarwalVZSGGB24}, respectively. DAgger and OPD achieve significantly better performance and much lower regret than BC and naive knowledge distillation, as the state distribution gap has been largely bridged, but at the cost of coupling inference and optimization, much like RL. Nonetheless, OPD inspired Multi-teacher OPD (MOPD)~\cite{DBLP:journals/corr/abs-2601-02780,ma2026mopd} to become the state-of-the-art for training LLMs across vast domains.

\subsection{Offline Reinforcement Learning}
Addressing the distribution mismatch between training trajectories and the policy being optimized is a long-standing topic in reinforcement learning. Offline RL, in particular, must address the most significant distribution gap, where static trajectories generated by any behavior policy may be used to optimize the performance of a completely unrelated policy~\cite{DBLP:journals/corr/abs-2005-01643}.
Distribution mismatch can cause convergence issues~\cite{baird1995residual} and value estimation biases~\cite{DBLP:conf/icml/FujimotoMP19}, and deteriorate or completely ruin the performances of existing RL algorithms.
Typical methods for addressing this challenge include explicit policy regularization to restrict the deviation of the policy distribution~\cite{DBLP:conf/icml/FujimotoMP19,DBLP:conf/nips/KumarFSTL19}, conservative value estimation to correct value overestimation~\cite{DBLP:conf/nips/KumarZTL20}, and importance-sampling-based off-policy correction to compensate for the deviation in the trajectory distribution.
The replay memory, or replay buffer, has been widely applied in off-policy RL algorithms to enhance data reuse and improve sample efficiency~\cite{lange2010deep, mnih2015human, ritter2026llms, arnal2026efficient}. However, most methods retain only a time window of historical trajectories to avoid a large distribution gap. The application of replay memories from entirely different models remains underexplored, especially in the context of training LLMs.

\section{Methodology}
\subsection{Preliminaries}
\label{sec:prelim}

Consider $x$ to be the context, which includes the user prompt and dialogue history, and $y$ to be the answer output of the LLM. We use the policy $\pi(\cdot|x)$ to represent the distribution of the LLM's generation. We define each task domain as $\tau_d = (X_d = \{x^{(i)}_d\}, r_d(\cdot))$, where $X_d$ is a collection of tasks or user queries, $r_d(\cdot)$ is the evaluation function for domain $\tau_d$, which can potentially be a reward model or rules. Given $D$ domains $\tau_1, \dots, \tau_D$, a LLM $\pi$ is first trained independently in each domain as follows:
\begin{equation}
     \pi \xrightarrow{\text{RL}(X_d, r_d)} \pi^{*}_d,
\end{equation}
with $\pi^{*}_d$ representing the teacher policy for domain $d$. However, $\pi^{*}_d$ is a domain specialist, as it typically fails in different domains. The most straightforward approach to distilling the teachers is behavior cloning (BC)~\cite{DBLP:conf/nips/Pomerleau88}, which forces each teacher to perform inference at its own mastered domain $y^*_d \sim\pi^*_d(\cdot | x_d)$, and minimizes the following imitation loss for the policy $\pi_{\theta}$ to be optimized:
\begin{equation}
\label{eq:bc}
\mathcal{L}_{\text{BC}}(\theta) \;\propto\;
-\sum_{d\le D}\mathbb{E}_{x \sim X_d, y \sim \pi^*_d(\cdot|x)}
\!\left[\, \frac{1}{|y|}\sum_{t}
\log \pi_\theta(y_t \mid x, y_{<t}) \,\right].
\end{equation}

The risk of BC is that when the self-generated trajectory $y_{<t}$ deviates from the teacher generated trajectory $y^*_{d,<t}$, the remaining trajectories might enter a "unfamiliar zone" and thus lead to the explosion of prediction error. To avoid the impact of this distribution shift, OPD~\cite{DBLP:conf/iclr/AgarwalVZSGGB24} utilizes supervision on self-generated trajectory instead of teacher generated trajectory. The loss function takes the following form:
\begin{align}
\label{sec:prelim:opd}
\mathcal{L}_{\text{OPD}}(\theta) &\propto
- \sum_{d \le D} \mathbb{E}_{x \sim X_d,\, y \sim \pi_\theta(\cdot \mid x)}
\!\left[\, \frac{1}{|y|} \sum_{t}
A^{\text{OPD}}_{t}\;
\log \pi_\theta(y_t \mid x,y_{<t}) \,\right], \\
A^{\text{OPD}}_{t} &= \text{stop-gradient}\left[\log \frac{\pi_d^*(y_t|x,y_{<t})}{\pi_\theta(y_t|x,y_{<t})}\right]. \nonumber
\end{align}
where the KL divergence between the teacher and the student is minimized on the student-generated trajectories instead of the teacher's. The advantage $A^{\text{OPD}}_{t}$ penalizes tokens that are less likely to be generated by the teacher and more likely to be generated by the student, while encouraging the reverse.
MOPD~\cite{DBLP:journals/corr/abs-2601-02780,ma2026mopd} further notices the lag between the student policy that generates the rollouts and the current policy under optimization, and introduces an importance weight $w_t$ to re-weight the advantage. It also introduces a reward-based advantage (such as the GRPO advantage) in addition to the KL divergence, forming a hybrid method integrating RL and policy distillation, which gives the advantage value of the following final form, with $[\,\cdot\,]_{a}^{b}$ representing clipping the importance ratio to $[a,b]$.
\begin{align}
\label{eq:mopd-adv}
A^{\text{MOPD}}_{t} &= w_t (A^{\text{OPD}}_{t} + \alpha A_t^{\text{GRPO}}),\\
& y \sim \pi_{\theta_\text{lag}}(\cdot|x), w_t=\text{stop-gradient}\left[\left[\frac{\pi_{\theta}(y_t \mid x,y_{<t})}{\pi_{\theta_{\text{lag}}}(y_t \mid x,y_{<t})}\right]_a^b\right] \nonumber
\end{align}

\subsection{Replay-recycling for Expert-to-Generalist Distillation with Offline RL}
\label{sec:method}

Instead of utilizing the teacher's policy $\pi^*_d$ to supervise the trajectory, we choose to recycle the experience within the RL training process of each teacher $\text{RL}(X_d,r_d)$, which we denote as $\mathcal{E}_d=\{(x^{(i)},y^{(i)},r^{(i)})\}$. The motivation for recycling the teacher's experience is straightforward: since domain experience drives the model to become domain specialists, accumulating experiences across those domains should at least be capable of driving the model to become competent in those domains. The only problem is the distribution shift; therefore, it is desirable to apply off-policy correction and offline learning algorithms. We name this method Replay-recycling for expert-to-generalist distillation with offline RL, or REGEN for short. The target loss of REGEN is motivated by two key considerations:
\begin{itemize}
    \item Existing Offline RL methods without importance sampling such as DPO~\cite{DBLP:conf/nips/RafailovSMMEF23} render most offline samples ineffective, while off-policy GRPO~\cite{mroueh2025revisiting} introduces importance sampling but employs symmetric clipping designed for near on-policy updates, zeroing out the gradient of any sequence whose importance ratio falls outside the given clipping interval and thereby discarding most learning signals under the large distribution shift of accumulated experiences. In contrast, TOPR~\cite{DBLP:journals/corr/abs-2503-14286} adopts Truncated Importance Sampling (TIS), which clips only the importance weight to $[0,1]$ that modulates the gradient while leaving the underlying policy gradient term intact.
    \item Concerning unified cross-domain optimization, the original reward signal employed by TOPR can be adversely affected by discrepancies in reward settings across different domains. A more principled alternative is to replace the raw reward with a normalized advantage following GRPO. However, GRPO normalizes the advantage exclusively within rollouts generated consecutively by a unique policy, here we need to compute the advantage over the history, with different rollouts possibly coming from different policy snapshots.
\end{itemize} 
Consequently, REGEN adopts the following loss function for offline trajectories:
\begin{align}
\label{eq:regen-obj}
\mathcal{L}_{\text{REGEN}}(\theta) \;&=\;
- \sum_{d \le D} \mathbb{E}_{(x,y,r) \sim \mathcal{E}_d} \left[ A^{\text{REGEN}} \cdot \frac{1}{|y|} \sum_t \log \pi_{\theta}(y_t|x,y_{<t}) \right], \\
A^{\text{REGEN}}&=\begin{cases}
\hat{A}, & \text{if } r \geq r_d^H \\
\text{stop-gradient}\!\left[\,\left[\dfrac{\pi_{\theta}(y|x)}{\mu(y|x)}\right]_0^1\,\right] \hat{A}, & \text{else} \nonumber
\end{cases},\\
\hat{A}&=\dfrac{r-\hat{\mathbb{E}}(r|x,\mathcal{E}_d)}{\sqrt{\hat{\mathbb{V}}(r|x,\mathcal{E}_d)}+\varepsilon}.
\end{align}
$r_d^H$ is a reward threshold defined per domain, serving as the criterion for classifying a sample as positive or negative. $\hat{\mathbb{E}}$ and $\hat{\mathbb{V}}$ are the empirical mean and variance, respectively. Notice that $\mu$ here is the policy that was used to sample the rollout $y$ at that moment. Therefore, it does not refer to a single uniform policy, but rather a collection of policies accumulated across the training process and across different domains. Another issue is that for each $x \in X_d$, the corresponding output trajectories recorded in $\mathcal{E}_d$ may be generated by models of various training stages and can become excessively large in number. To address these challenges, REGEN introduces two additional techniques: augmented experience recording and class-balanced sampling.

\paragraph{Augmented experience recording.} This phase harvests all rollouts from each teacher's online RL training without forgetting. To ensure $\mu(y|x)$ in Equation~\ref{eq:regen-obj} can be effectively retrieved, we record the Negative Log Likelihood (NLL) $z= - \sum_t \log \mu(y_t|x,y_{<t})$ as the policy generating the rollout, and therefore the replay buffer is augmented by the NLL and represented as:
\begin{equation}
\label{eq:replay-tuple}
\Xi_d \;=\; \{(\, x^{(i)},y^{(i)},z^{(i)},r^{(i)}) \},
\end{equation}
this replay buffer then ensures the estimation of $\mu$ can be efficiently recovered by $\mu(y^{(i)}|x^{(i)})=\exp (-z^{(i)})$.

\paragraph{Class-balanced sampling.}
Similar to GRPO, we group the responses in the replay buffer by the same query. Since an overwhelming number of negative samples has been found to be harmful to RL training~\cite{DBLP:conf/nips/HongKKBSPLGA23}, while uniform sampling---in accumulated rollouts where positives dominate---repeatedly draws highly similar positive samples, reducing sample diversity, collapsing the advantage signal and driving the model to overfit the concentrated positive mode, and positives-only filtering discards negatives entirely, losing the contrastive signal needed to localize decision boundaries~\cite{DBLP:journals/corr/abs-2503-14286}, we additionally apply Class-Balanced Sampling (CBS)~\cite{cui2019class} to adjust the distribution between positive and negative samples. Specifically, we retain only the \emph{valid} queries, i.e., those that have at least one positive response and one negative response. For each query $x$, define the positive and negative groups:
\begin{equation}
\label{eq:cbs_1}
\Xi_d^\pm(x) = \{(x,y,z,r)\in\Xi_d : r \gtreqless r_d^H\},
\end{equation}
where $\gtreqless$ is $\ge$ for $+$ and $<$ for $-$.
To balance them, we downsample each group with probability
\begin{equation}
\label{eq:cbs_2}
p^\pm(x) = \frac{\min(|\Xi_d^+(x)|,|\Xi_d^-(x)|)}{|\Xi_d^\pm(x)|},
\end{equation}
and form $\mathcal{E}_d(x) = \text{Down-Sample}(\Xi_d^+(x),p^+(x)) \cup \text{Down-Sample}(\Xi_d^-(x),p^-(x))$.
Offline training is based on the down-sampled trajectories over all queries $\mathcal{E}_d = \{\mathcal{E}_d(x) : x\in X_d\}$. Notice that in Equation~\ref{eq:regen-obj}, the advantage $\hat{A}$ is calculated over the down-sampled group $\mathcal{E}_d(x)$ with respect to each query $x$. A review of the REGEN method is shown in Algorithm~\ref{alg:regen}.

\begin{algorithm}
\caption{Replay-recycling for Expert-to-Generalist Distillation with Offline RL (REGEN)}
\label{alg:regen}
\begin{algorithmic}[1]
\Require Tasks from $D$ domains: $\{(X_d,r_d)\}$; Initial model: $\pi$
\State \textbf{Experience Recording}: Independently train teachers and collect experiences following Equation~\ref{eq:replay-tuple}.
\State \textbf{Class-balanced Sampling}: Filter and down-sample the replay buffer according to Equations~\ref{eq:cbs_1} and~\ref{eq:cbs_2}.
\State \textbf{Off-Policy Optimization}: Update the student model using the REGEN objective functions (Equation~\ref{eq:regen-obj}).
\State \Return The generalist policy $\pi_{\theta}$.
\end{algorithmic}
\end{algorithm}

Note that REGEN can be readily extended to multi-turn tasks, where $x$ represents both the query and the dialogue context at each turn. Moreover, it does not require the rollout-generating policy and the policy under optimization to share the same backbone or the same vocabulary as the teacher. These features endow the method with exceptional flexibility in data accumulation and offline training.

\section{Experiments}

\subsection{Experiment Settings}
\label{sec:exp:setup}

To validate the effectiveness of REGEN in training generalist models, we perform joint training across three domains using the Qwen2.5-1.5B-Instruct~\cite{DBLP:journals/corr/abs-2412-15115} as the base model: mathematical problem solving (\textbf{Math}), code generation (\textbf{Code}), and instruction following (\textbf{Alignment}). Our training corpora consist of GSM8K~\cite{DBLP:journals/corr/abs-2110-14168} and MATH~\cite{DBLP:conf/nips/HendrycksBKABTS21} for Math, KodCode-Light-RL-10K~\cite{DBLP:conf/acl/XuLYZP25} for Code, and an IFEval-style instruction-following dataset~\cite{huggingfaceh4_ifevallike} for Alignment. For evaluation, we primarily report inference-time decoding accuracies on GSM8K, MATH, HumanEval~\cite{DBLP:journals/corr/abs-2107-03374}, and MBPP~\cite{DBLP:journals/corr/abs-2108-07732}; for IFEval~\cite{DBLP:journals/corr/abs-2311-07911}, we report both loose-match and strict-match accuracies.

We first train domain-expert teachers via online RL (also using the Qwen2.5-1.5B-instruct as the base model), specifically GRPO~\cite{DBLP:journals/corr/abs-2402-03300}, under a unified configuration shared across domains. The specific reward allocation modes for different domains are explained in Appendix~\ref{sec:reward_design}. We set the maximum generation length to $1024$, sample $n=4$ responses per prompt, use a global batch size of $128$, and adopt a learning rate of $1 \times 10^{-6}$. Each RL run is trained for $4$ epochs. Overall, training these teachers yields approximately $800K$ logged trajectories in total and produces one expert model per domain. Based on the trained teachers and logged trajectories, we mainly compare the following methods:

\begin{itemize}
    \item \emph{Behavior Cloning} (BC)~\cite{DBLP:conf/nips/Pomerleau88}: We select the best checkpoint from each online RL run as the domain-expert teacher and employ it to re-label answers to the training queries across all three domains. The student base model is then directly trained on this re-labeled corpus using Equation~\ref{eq:bc}.
    \item \emph{Multi-Teacher On-Policy Distillation} (MOPD)~\cite{DBLP:journals/corr/abs-2601-02780,ma2026mopd}: After acquiring the domain-expert teachers, for each training query across all three domains, we generate a new rollout using the policy under optimization, evaluate the answer with a reward, and relabel each token according to the corresponding domain expert. The student model is then trained with Equation~\ref{eq:mopd-adv}.
    \item \emph{REGEN} : We apply CBS~\cite{cui2019class}(The reward threshold $r_d^H$ for each domain is $+1$) directly to the logged trajectories, yielding $140K$ trajectories with an equal number of positive and negative examples. We then train the student base model on this sampled dataset by minimizing the REGEN objective in Equation~\ref{eq:regen-obj}.
\end{itemize}

For training BC, MOPD, and REGEN, we adopt experimental settings similar to those for teacher training, with a maximum sequence length of $1024$, a batch size of $128$, a slightly lower learning rate of $5.0 \times 10^{-7}$, and $4$ training epochs. To ensure a fair comparison—even though REGEN can leverage a much larger corpus at minimal additional cost—we keep the total corpus used across the three methods to an identical size of $140K$, distributed roughly uniformly among the domains. Specific distribution statistics of the collected replay memory are provided in Appendix~\ref{sec:dist_stats}.
All experiments are conducted on a single node equipped with 4 NVIDIA L40S GPUs, each with 48\,GB of memory. Each device uses a maximum local batch size of $4$, and we apply $8$-step gradient accumulation to reach an effective global batch size of $128$.

\subsection{Main Results} 
\label{sec:exp:main}

Table~\ref{tab:main-acc} summarizes the performance of different methods across five benchmarks, reported under multiple evaluation metrics. Overall, REGEN consistently matches or surpasses MOPD on most metrics, while substantially outperforming BC, particularly on Code and Alignment. For the 1.5B student, REGEN achieves the best results on MATH (pass@1 and maj@4), HumanEval (pass@10), and MBPP (pass@10) compared to BC and MOPD. For the 0.5B student, REGEN leads on all benchmarks (except GSM8K and IFEval) under multiple metrics, demonstrating its robustness across model scales. Notably, REGEN achieves this competitive accuracy with considerably higher training efficiency than MOPD. Since the student trainer maintains only $\pi_\theta$, without hosting either the teacher model or a rollout engine during training, REGEN eliminates the online teacher inference that dominates MOPD's computational cost, thereby achieving substantially higher training throughput and lower per-token latency, as shown in Table~\ref{tab:training_efficiency}.

Despite these gains, on IFEval, REGEN slightly lags MOPD under loose/strict metrics (e.g., -2.8 on loose metric for 0.5B, -3.5 on strict metric for 1.5B) due to coarser trajectory-level supervision, whereas MOPD benefits from dense token-level KL guidance from online teacher rollouts. In addition, both REGEN and MOPD remain noticeably behind the code teacher on HumanEval under pass@1. We attribute this gap to a fundamental limitation of policy distillation under distribution shift~\cite{DBLP:journals/jmlr/RossB10}. Specifically, the training corpus KodCode-Light-RL-10K consists primarily of code generation tasks similar to MBPP, whereas HumanEval evaluates code completion. Since both methods distill the teacher's behavior only on the training distribution, the mismatch between generation and completion prevents full inheritance of the teacher's capability on the target benchmark.


\begin{table}[htbp]
\footnotesize
\centering
\caption{Accuracy comparison across benchmarks and multiple evaluation metrics. Teacher is the same online-RL expert (using the Qwen2.5-1.5B-instruct as the base model). For each student model and each metric column, the best among BC, MOPD, and REGEN is bolded. Superscripts in the REGEN rows indicate the difference relative to the MOPD value in the same column.}
\label{tab:main-acc}
\begin{tabular}{l|cccccccccc}
\toprule
\multirow{2}{*}{Method} & \multicolumn{2}{c}{GSM8K} & \multicolumn{2}{c}{MATH} & \multicolumn{2}{c}{HumanEval} & \multicolumn{2}{c}{MBPP} & \multicolumn{2}{c}{IFEval} \\
\cmidrule(lr){2-3} \cmidrule(lr){4-5} \cmidrule(lr){6-7} \cmidrule(lr){8-9} \cmidrule(lr){10-11}
 & pass@1 & maj@4 & pass@1 & maj@4 & pass@1 & pass@10 & pass@1 & pass@10 & loose & strict \\
Teacher & 81.2 & 83.1 & 57.0 & 59.6 & 66.5 & 70.1 & 70.4 & 76.7 & 73.0 & 65.2 \\
\midrule
\multicolumn{11}{l}{\textit{Student: Qwen2.5-1.5B-Instruct}} \\
Base & 76.6 & 77.3 & 53.6 & 57.8 & 55.5 & 73.8 & 65.3 & 84.9 & 52.7 & 49.7 \\
\rowcolor{highlight}
BC   & 79.4 & 81.3 & 57.0 & 60.8 & 57.3 & 68.9 & 66.9 & 81.7 & 65.4 & 60.1 \\
\rowcolor{highlight}
MOPD & \textbf{79.9} & 81.9 & 57.3 & 59.0 & 62.2 & 65.9 & \textbf{71.4} & 77.2 & \textbf{73.2} & \textbf{67.8} \\
\rowcolor{highlight}
REGEN& 79.5\textsuperscript{\tiny -0.4} & \textbf{82.6}\textsuperscript{\tiny +0.7} & \textbf{58.4}\textsuperscript{\tiny +1.1} & \textbf{61.4}\textsuperscript{\tiny +2.4} & \textbf{62.2}\textsuperscript{\tiny +0.0} & \textbf{70.1}\textsuperscript{\tiny +4.2} & 69.3\textsuperscript{\tiny -2.1} & \textbf{77.8}\textsuperscript{\tiny +0.6} & 71.5\textsuperscript{\tiny -1.7} & 64.3\textsuperscript{\tiny -3.5} \\
\midrule
\multicolumn{11}{l}{\textit{Student: Qwen2.5-0.5B-Instruct}} \\
Base & 48.7 & 52.1 & 32.6 & 35.5 & 39.0 & 48.8 & 48.4 & 60.3 & 38.1 & 36.2 \\
\rowcolor{highlight}
BC   & 51.9 & 55.5 & 35.2 & 37.5 & 40.6 & 45.7 & 47.1 & 57.7 & 46.2 & 44.0 \\
\rowcolor{highlight}
MOPD & \textbf{60.6} & \textbf{63.8} & 36.5 & 38.8 & 40.2 & 45.7 & 51.9 & 54.8 & \textbf{59.9} & \textbf{55.8} \\
\rowcolor{highlight}
REGEN& 57.8\textsuperscript{\tiny -2.8} & 59.0\textsuperscript{\tiny -4.8} & \textbf{38.1}\textsuperscript{\tiny +1.6} & \textbf{39.8}\textsuperscript{\tiny +1.0} & \textbf{41.5}\textsuperscript{\tiny +1.3} & \textbf{48.8}\textsuperscript{\tiny +3.1} & \textbf{52.6}\textsuperscript{\tiny +0.7} & \textbf{63.5}\textsuperscript{\tiny +8.7} & 57.1\textsuperscript{\tiny -2.8} & 52.7\textsuperscript{\tiny -3.1} \\
\bottomrule
\end{tabular}
\end{table}



\begin{table}[t]
\centering
\caption{Training efficiency of REGEN versus MOPD on four 48 GB NVIDIA L40S GPUs. MOPD places each of the three teachers on one GPU and the student on the remaining one, while REGEN trains the student on all four GPUs without any teacher serving.}
\label{tab:training_efficiency}
\footnotesize
\renewcommand{\arraystretch}{1.18}
\setlength{\tabcolsep}{4.5pt}

\begin{tabular}{lcc|cccc}
\toprule
& \multicolumn{2}{c|}{\textbf{Training Throughput}}
& \multicolumn{3}{c}{\textbf{Per-1k-Token Latency}}
& \textbf{Overall} \\
\cmidrule(lr){2-3}\cmidrule(lr){4-6}
\textbf{Method}
& \textbf{Tokens/s}
& \textbf{Speedup}
& \textbf{Forward}
& \textbf{Backward}
& \textbf{Total}
& \textbf{Speedup} \\
\midrule
MOPD
& 1,528.0
& 1.00$\times$
& 463.3
& 218.7
& 682.0
& 1.00$\times$ \\

\rowcolor{blue!8}
\textbf{REGEN (Ours)}
& \textbf{7,714.5}
& \textbf{5.05$\times$}
& \textbf{43.2}
& \textbf{86.4}
& \textbf{129.6}
& \textbf{5.26$\times$} \\
\bottomrule
\end{tabular}
\end{table}

\subsection{Ablations and Analysis}
\label{sec:exp:ablation}

\begin{figure}[t]
    \centering
    \includegraphics[width=0.8\linewidth]{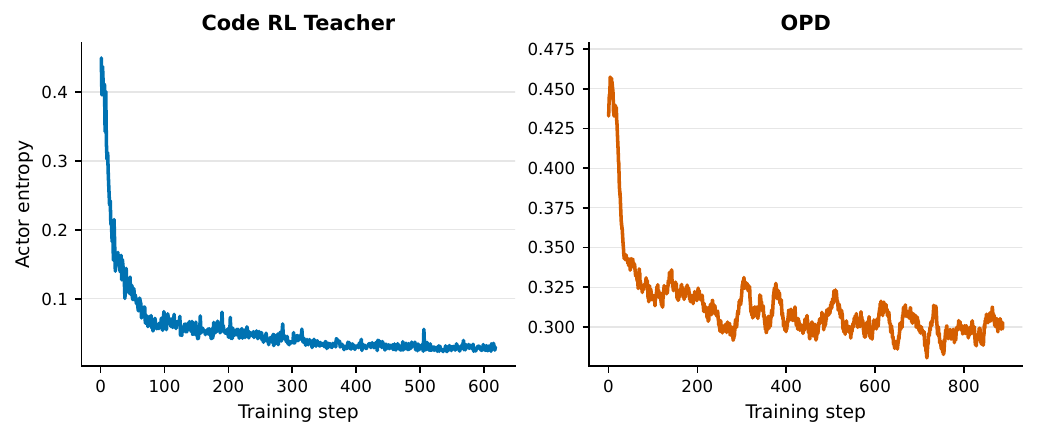}
    \caption{Training trajectories of the actor entropy for Code RL Teacher and OPD. The left and right panels show the corresponding entropy evolution over training steps, respectively.}
    \label{fig:actor-entropy}
\end{figure}

\paragraph{RQ1: How does REGEN compare to OPD in diversity exploration under pass@k when Teacher fails?}
The main experiments~\ref{sec:exp:main} reveal an interesting empirical phenomenon: on the two code generation tasks, HumanEval and MBPP, the Teacher fails to preserve its advantage under pass@k and can even fall below the original base model. We mainly attribute this to entropy collapse in the late stage of RL training, which limits the Teacher's diverse exploration. This limitation is intrinsic to online RL-style optimization and therefore directly affects OPD, since OPD continues to train the student through online rollouts and thus tends to inherit the same entropy-collapse behavior. For REGEN, however, the effect is only indirect: entropy collapse in the Code RL Teacher may produce many duplicated responses in the replay memory, and such duplicates can induce overfitting and degrade offline training. To investigate this, we use DeepSeek-V4 to remove duplicated responses, obtaining a $10K$ code training set, and retrain both REGEN and OPD on it (using Qwen2.5-1.5B-Instruct as the student model). Figure~\ref{fig:actor-entropy} shows the entropy curves of the Code RL Teacher and the newly trained OPD actor; OPD still exhibits entropy collapse. We regard this as an inherent limitation of RL training: since OPD is essentially RL, it naturally inherits this limitation, whereas REGEN is an offline method and is therefore not directly subject to entropy collapse. Existing work~\cite{chen2026does,yu2026dapo} strongly supports this argument. As shown in Table~\ref{tab:passk-ablation}, after removing duplicated data, REGEN significantly outperforms OPD under pass@k. Together with the main results in Section~\ref{sec:exp:main}, where the 0.5B model achieves pass@k no lower than the base model and higher than MOPD on HumanEval and MBPP, we tentatively conclude that REGEN is superior to OPD in diversity exploration.

\begin{table}[htbp]
\footnotesize
\centering
\caption{Pass@1 and pass@10 comparison of REGEN and OPD on HumanEval and MBPP under a failed Teacher. Results are obtained on a $10K$ code training set with duplicated responses removed, using Qwen2.5-1.5B-Instruct as the student model. Teacher selects the same code teacher as the main experiment. For each metric column, the best among Base, OPD, and REGEN is bolded.}
\label{tab:passk-ablation}
\begin{tabular}{l|cccc}
\toprule
\multirow{2}{*}{Method} & \multicolumn{2}{c}{HumanEval} & \multicolumn{2}{c}{MBPP} \\
\cmidrule(lr){2-3} \cmidrule(lr){4-5}
 & pass@1 & pass@10 & pass@1 & pass@10 \\
Teacher & 66.5 & 70.1 & 70.4 & 76.7 \\
\midrule
\multicolumn{5}{l}{\textit{Student: Qwen2.5-1.5B-Instruct}} \\
\rowcolor{highlight}
Base & 55.5 & 73.8 & 65.3 & 84.9 \\
\rowcolor{highlight}
OPD & \textbf{60.4} & 67.6 & \textbf{70.1} & 79.2 \\
\rowcolor{highlight}
REGEN& 59.1\textsuperscript{\tiny-1.3} & \textbf{75.6}\textsuperscript{\tiny+8.0 } & 69.3\textsuperscript{\tiny-0.8} & \textbf{88.3}\textsuperscript{\tiny+9.1} \\
\bottomrule
\end{tabular}
\end{table}

\paragraph{RQ2: How do the sampling strategy and training objective affect model performance?}

We compare all combinations of sampling strategy and training objective in Table~\ref{tab:grid_full}. Overall, the combination of CBS and REGEN achieves the best performance on almost all benchmarks. This confirms the effectiveness of CBS in correcting the imbalance inherent in accumulated rollouts and demonstrates that the asymmetric TIS objective in REGEN handles distribution shift more effectively than symmetric clipping and preference-based alternatives.
The only exception is MBPP, where positives-only SFT achieves the best performance. This difference may stem from the nature of code generation: a program forms an integrated computation through its control flow, so an error typically propagates globally rather than remaining localized. Consequently, a fully correct program already serves as a clean imitation target, whereas an incorrect one provides little transferable supervision. By contrast, in mathematical reasoning, positive and negative solutions usually share most intermediate reasoning steps and diverge only at the erroneous step, while in IFEval the soft compositional constraints preclude a simple binary correctness criterion. In both settings, contrasting a correct solution with a near-miss negative provides critical supervision for localizing the decision boundary.

\begin{table}[t]
\centering
\caption{Ablation of the sampling strategy and training objective on Qwen2.5-1.5B-Instruct, using the accumulated rollout trajectories logged along the online teachers' RL runs. The best result in each column is highlighted in bold. GRPO denotes off-policy GRPO.}
\label{tab:grid_full}
\footnotesize
\renewcommand{\arraystretch}{1.15}
\setlength{\tabcolsep}{4.2pt}

\begin{tabular}{llcccccc}
\toprule
\textbf{Sampling}
& \textbf{Objective}
& \textbf{GSM8K}
& \textbf{MATH}
& \textbf{HumanEval}
& \textbf{MBPP}
& \textbf{IFEval}
& \textbf{Avg.} \\
\midrule

\textbf{Positives Only}
& \textbf{SFT}
& 77.7 & 57.3 & 61.0 & \textbf{72.0} & 68.2 & 67.2 \\
\midrule

\multirow{3}{*}{\textbf{Uniform}}
& \textbf{TOPR}
& 77.7 & 56.7 & 59.1 & 69.6 & 70.1 & 66.6 \\
& \textbf{GRPO}
& 74.9 & 54.1 & 59.1 & 66.9 & 67.8 & 64.6 \\
\rowcolor{blue!8}
& \textbf{REGEN}
& 77.2 & 56.7 & 60.4 & 68.5 & 71.0 & 66.8 \\
\midrule

\multirow{4}{*}{\textbf{CBS}}
& \textbf{TOPR}
& 78.5 & 57.6 & 61.6 & 70.1 & 66.7 & 66.9 \\
& \textbf{GRPO}
& 76.5 & 52.6 & 60.4 & 68.8 & 60.0 & 63.7 \\
& \textbf{DPO}
& 75.3 & 54.7 & 60.4 & 65.9 & 56.9 & 62.6 \\
\rowcolor{blue!8}
& \textbf{REGEN}
& \textbf{79.5} & \textbf{58.4} & \textbf{62.2}
& 69.3 & \textbf{71.5} & \textbf{68.2} \\
\bottomrule
\end{tabular}
\end{table}

\paragraph{RQ 3: How does the performance of REGEN differ between multi-domain and single-domain training?}
We investigate whether training a single student jointly across all domains incurs any performance degradation compared with training a dedicated specialist for each domain. Table~\ref{tab:single_multi_best} compares our multi-domain offline student with single-domain students trained using the same objective on each domain independently. Across all five benchmarks, the performance gap never exceeds $2$ accuracy points, and neither setting consistently outperforms the other. These results indicate that aggregating data from multiple domains introduces little to no cross-domain interference, demonstrating that REGEN remains stable and effective when a single student is shared across diverse domains.

\begin{table}[t]
\centering
\caption{Multi-domain versus single-domain training with REGEN on
Qwen2.5-1.5B-Instruct. \emph{Multi} trains one student jointly over all three
domains, whereas \emph{Single} trains a separate student per domain. The two
settings share identical training hyperparameters.The gap is computed as Single-domain minus Multi-domain.}
\label{tab:single_multi_best}
\footnotesize
\renewcommand{\arraystretch}{1.15}
\setlength{\tabcolsep}{5pt}

\begin{tabular}{lccccc}
\toprule
\textbf{Training Setting}
& \textbf{GSM8K}
& \textbf{MATH}
& \textbf{HumanEval}
& \textbf{MBPP}
& \textbf{IFEval} \\
\midrule
Multi-domain
& 79.5
& 58.4
& 62.2
& \textbf{69.3}
& 71.5 \\
Single-domain
& \textbf{80.2}
& \textbf{58.3}
& \textbf{63.4}
& 69.0
& \textbf{73.2} \\
\midrule
\rowcolor{blue!8}
Gap
& +0.7
& -0.1
& +1.2
& -0.3
& +1.7 \\
\bottomrule
\end{tabular}
\end{table}



\section{Conclusion and Discussion}
\label{sec:conclusion}
We propose REGEN, a method that recycles the replay memory accumulated during RL training of domain-expert teachers to subsequently train a generalist student. Through class-balanced sampling and an advantage-weighted asymmetric offline objective, REGEN achieves accuracy comparable to MOPD at substantially lower computational cost, surpassing naive behavior cloning by a significant margin. More broadly, REGEN reframes any RL run as a recyclable data-production process rather than a one-off learning stage, offering a scalable path toward large-scale post-training.

Our work effectively demonstrates the potential of reusing replay memory; however, the optimal strategy for leveraging this resource remains an open question worthy of deeper investigation. For example, are the negative samples discarded by CBS truly devoid of value, or could more sophisticated reuse methods unlock additional utility? Furthermore, while REGEN currently treats all replay data uniformly regardless of their generation stage, incorporating stage-specific information may enable more effective utilization. We leave these directions to future work.

\bibliographystyle{unsrtnat}
\bibliography{reference}

\appendix

\section{Reward Design Across Domains}
\label{sec:reward_design}

The three task domains adopt distinct reward schemes tailored to their evaluation characteristics.

For the \textbf{Math} domain, the reward is binary. Both GSM8K and MATH employ prompts that constrain the final answer within a \texttt{box\{\}} format, and the reward is determined by matching the extracted answer against the ground truth. During the online RL training of each Math teacher, we assign a reward of $1$ for correct answers and $0$ for incorrect ones, as negative rewards would dilute the learning signal and hinder teacher training. However, in the TOPR baseline, where trajectories are weighted by their rewards instead of advantages, a reward of $0$ for negative samples would cause them to be entirely discarded. To address this, we redefine the trajectory-level reward of negative samples as $-1$ during TOPR training, which aligns with the original TOPR design and ensures that negative samples contribute a meaningful contrastive signal.

For the \textbf{Code} domain, the reward is continuous in $[0, 1]$, reflecting the fraction of passed test cases. A program that passes all tests receives a reward of $1$, while a completely failing program receives $0$, with intermediate values proportional to the test pass rate.
For the \textbf{Alignment} domain, the reward is also continuous in $[0, 1]$, representing the proportion of verifiable constraints satisfied by the generated response. This fine-grained reward provides nuanced supervision compared to the binary Math domain.

\section{Distribution Statistics of Replay Memory}
\label{sec:dist_stats}

\begin{figure}[htbp]
    \centering
    \includegraphics[width=\linewidth]{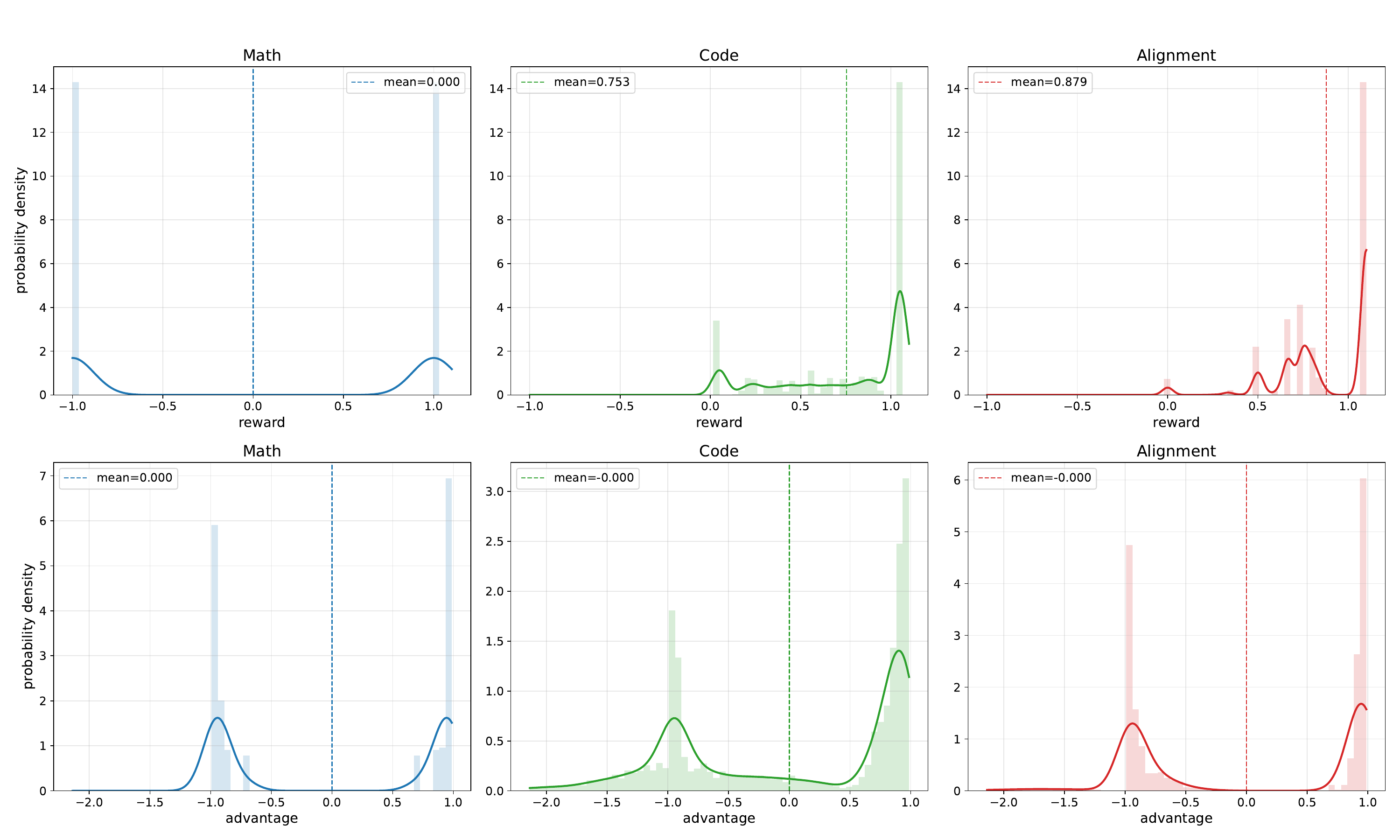}
    \caption{Reward (Top) vs Advantage (Bottom) Distribution by Domain. The dashed lines indicate the mean values.}
    \label{fig:reward_advantage_dist}
\end{figure}

Figure~\ref{fig:reward_advantage_dist} illustrates the reward and advantage distributions of the accumulated replay memory across the three domains. As expected from the reward design described in Appendix~\ref{sec:reward_design}, the Math domain exhibits a bimodal reward distribution concentrated at $-1.0$ and $1.0$, while Code and Alignment display continuous distributions over $[0, 1]$. After group-relative normalization, the advantage distributions of Code and Alignment are centered around $0.0$ with a continuous range, whereas the Math advantage distribution mirrors its binary reward structure. Despite the heterogeneity in reward scales, the advantage normalization in REGEN provides a unified, scale-invariant learning signal across all domains.

\end{document}